\documentclass[conference]{IEEEtran}

\IEEEoverridecommandlockouts
\usepackage{cite}
\usepackage{amsmath,amssymb,amsfonts}
\usepackage{algorithmic}
\usepackage{graphicx}
\usepackage{textcomp}
\usepackage{xcolor}
\usepackage[margin=5pt]{subfig}
\usepackage{multirow}
\def\BibTeX{{\rm B\kern-.05em{\sc i\kern-.025em b}\kern-.08em
    T\kern-.1667em\lower.7ex\hbox{E}\kern-.125emX}}

\begin{document}

\title{Interpreting Deep Models \\through the Lens of Data}

\author{\IEEEauthorblockN{Dominique Mercier\IEEEauthorrefmark{1}\IEEEauthorrefmark{2}, Shoaib Ahmed Siddiqui\IEEEauthorrefmark{1}\IEEEauthorrefmark{2}, Andreas Dengel\IEEEauthorrefmark{1}\IEEEauthorrefmark{2}, Sheraz Ahmed\IEEEauthorrefmark{2}}
\IEEEauthorblockA{\IEEEauthorrefmark{1}Technische Universität Kaiserslautern, Germany.}
\IEEEauthorblockA{\IEEEauthorrefmark{2}DFKI GmbH, Germany.\\EMail: \{dominique.mercier, shoaib\_ahmed.siddiqui, andreas.dengel, sheraz.ahmed\}@dfki.de}}

\maketitle

\begin{abstract}
Identification of input data points relevant for the classifier (i.e. serve as the support vector) has recently spurred the interest of researchers for both interpretability as well as dataset debugging. This paper presents an in-depth analysis of the methods which attempt to identify the influence of these data points on the resulting classifier. To quantify the quality of the influence, we curated a set of experiments where we debugged and pruned the dataset based on the influence information obtained from different methods. To do so, we provided the classifier with mislabeled examples that hampered the overall performance. Since the classifier is a combination of both the data and the model, therefore, it is essential to also analyze these influences for the interpretability of deep learning models. Analysis of the results shows that some interpretability methods can detect mislabels better than using a random approach, however, contrary to the claim of these methods, the sample selection based on the training loss showed a superior performance.
\end{abstract}


\maketitle

\let\svthefootnote\thefootnote
\let\thefootnote\relax\footnote{\copyright 2020 IEEE. Personal use of this material is permitted. Permission from IEEE must be obtained for all other uses, in any current or future media, including reprinting/republishing this material for advertising or promotional purposes, creating new collective works, for resale or redistribution to servers or lists, or reuse of any copyrighted component of this work in other works.} 
\addtocounter{footnote}{-1}\let\thefootnote\svthefootnote

\section{Introduction}
Deep learning methods are currently at the forefront of technology and have been employed in many diverse domains such as image classification~\cite{cirecsan2012multi}, object segmentation~\cite{noh2015learning}, text classification~\cite{zhang2015character}, speech recognition~\cite{hinton2012deep}, and activity recognition~\cite{yang2015deep}. Deep learning is a subset of representation learning methods and therefore, it can automatically discover the relevant features for any given task. These methods rely on a large amount of data to achieve generalization and since the features are extracted from the data itself, this also raises implications on the quality of the dataset. As both the model and the data are important to create a good classifier, and an analysis of the influential data-points in addition to the commonly analyzed influence of the architecture~\cite{tsviz} is essential. 

In this paper, we explore this direction (i.e., using influence functions~\cite{koh2017understanding} and representer points~\cite{yeh2018representer}) as a way of interpreting a classifier and shade light on the underlying structure. In particular, we analyze the robustness properties of the classifier concerning the influential training points. Also, the analysis of results provides insights concerning the generalization capabilities of the classifiers. Finally, our experiments highlight the focus/attention of the classifiers by comparing the influence of the different training samples.

\section{Related Work}
Nowadays many datasets are publicly available but these datasets vary in their size and quality. They are most times used without any inspection because the manual inspection is not always feasible due to effort. Unfortunately, in some domains, it is crucial to have a very high-quality dataset as its a part of the model. Particularly, safety-critical application areas require explainable and reliable systems. This explainability needs to be fulfilled for the complete model including not only the prediction but furthermore the internal computations, structure decisions, and the data. Especially, most explainability papers focus to explain only the network and exclude the data. However, high-quality data is essential and there exist two categories of approaches for dataset debugging, namely traditional statistical methods and recently introduced interpretability methods.

\subsection{Traditional statistical dataset debugging}
To improve the dataset quality Zhu et al.~\cite{zhu2003eliminating} presented a rule-based approach to identify mislabeled instances in large datasets. Their approach partitions the data into smaller subsets and applies a rule set to evaluate the dataset and get information about the samples. Based on the above-mentioned approach, Guan et al.~\cite{guan2014detecting} evaluated the use of a multiple voting scheme, instead of the previously used single voting scheme, for mislabeling correction. In contrast to the rule-based approach, Sturm~\cite{sturm2012analysis} investigated the mislabel correction task by using a Bayesian classifier to correct the training data by predicting the probabilities for all data points belonging to all considered class labels. Another approach detecting data samples that are likely to be mislabeled was proposed by Akusok et al.~\cite{akusok2015md} assuming that the generalization error of a model decreases if a mislabeled sample is changed to the correct label. Facing the problem from a different perspective, Frénay and Verleysen~\cite{frenay2013classification} explain how to deal with label noise highlighting the importance of the problem, types of label noise, and different dataset cleansing methods. Finally, Patrini et al.~\cite{patrini2017making} presented an approach to correct the loss of a network concerning the probability of a label to be flipped using stochastic methods to compute the probability. 

\subsection{Interpretability based dataset debugging}
In contrast to the traditional statistical approaches Koh and Liang~\cite{koh2017understanding} and Yeh et al.~\cite{yeh2018representer} utilized the power of interpretability methods to identify possible mislabels. Therefore, Koh and Liang propose influence functions computed using the gradients to trace the influence of the samples for a given prediction enabling the separation into helpful and harmful samples used during dataset debugging. Precisely, they efficiently compute the influence by up-weighting a sample using the hessian. Conversely, Yeh et al. proposed to decompose the pre-activations resulting in a linear combination of activations of the training samples used as weights explaining the influence of the samples. To do so, they used the pre-activaitons and fed them to a solver using a self-defined loss.

\section{Datasets}
\begin{table}[!t]
\tiny
\renewcommand{\arraystretch}{1.3}
\caption{Dataset Properties.}
\label{tab:datasets}
\centering
\begin{tabular}{|c|c|c|c|c|c|c|}
    \hline
    \textbf{Dataset} & \textbf{Train} & \textbf{Validation} & \textbf{Test} & \textbf{Length} & 
    \textbf{Channels} & \textbf{Classes} \\
    \hline
    Synthetic Anomaly Detection & 45000 & 5000 & 10000 & 50 & 3 & 2 \\
    Character Trajectories & 1383 & 606 & 869 & 206 & 3 & 20 \\
    FordB & 2520 & 1091 & 810 & 500 & 1 & 2 \\
    \hline
\end{tabular}
\end{table}

Subjectivity and cherry-picking are two major challenges for explainability methods. To provide evidence for the methods and prove the correctness of the experiments it is important to conduct experiments using different datasets. Therefore, we used three different publicly available datasets including point anomaly, sequence anomaly, and a classification task. Precisely, we used the character trajectories and FordB dataset from the UCR Time Series Classification Repository\footnote{http://www.timeseriesclassification.com/} and a synthetic anomaly detection dataset~\cite{tsviz}. Furthermore, these datasets cover both binary and multi-class classification tasks and come with different sequence lengths and a different number of channels to achieve the largest possible variation of properties.

\section{Analysis and Discussion}
During our analyses, we conducted different experiments to shed light on several aspects concerning debugging rates, accuracy, time consumption, and interpretability. Besides a random selection used as a baseline and the network loss representing a direct measure, we used two well-known network interpretability methods that claim to improve mislabel correction namely influence functions~\cite{koh2017understanding} and representer points~\cite{yeh2018representer}. Finally, we compare the used methods and list their advantages and drawbacks. To create the datasets for the debugging, we flipped some labels within the dataset original datasets.

\subsection{Mislabel correction approaches}
In order to understand the debugging priority, we explain the ranking mechanisms excluding the random and loss approach as they are intuitive. Firstly, we used the influence functions~\cite{koh2017understanding} providing negative and positive values to highlight harmful and helpful samples. Therefore, we can inspect the most harmful, most helpful, and most influencing samples. In addition, we can compute the influence scores for each class individually (classwise) or over the complete set. Secondly, we use the representer values~\cite{yeh2018representer} that only provide information about inhibitory (low) and excitatory (high) samples.

\subsection{Experiment 1: Mislabel correction performance}
\begin{table*}[!t]
\tiny
\centering
\caption{Detected mislabeled in percentage sorted by different datasets, dataset qualities (percentage of mislabeled data), and inspection (percentage of inspected data).}
\label{tab:correction}
\begin{tabular}{|c|c|c|c||c|c|c|c|c|c||c|c||c||c|}
\hline
        \multirow{3}{*}{\textbf{Dataset}} & \multirow{3}{*}{\textbf{Model Acc.}} & \multirow{3}{*}{\textbf{Mislabeled}} & \multirow{3}{*}{\textbf{Inspected}} &
        \multicolumn{10}{c|}{\textbf{Detected Mislabels}} \\
        \cline{5-14}
        & & & & \multicolumn{6}{c||}{\textbf{Influence-based~\cite{koh2017understanding}}} & \multicolumn{2}{c||}{\textbf{Representer theorem~\cite{yeh2018representer}}} & \multirow{2}{*}{\textbf{Loss}} & \multirow{2}{*}{\textbf{Random}} \\
        & & & & \textbf{classwise low} & \textbf{classwise high} & \textbf{classwise absolute} & 
        \textbf{low} & \textbf{high} & \textbf{absolute} & \textbf{low} & \textbf{high} & & \\
        \hline
        \multirow{9}{*}{Anomaly} & \multirow{3}{*}{98.48\%} & \multirow{3}{*}{10\%} & 10\% & 
            14.34\% & 82.6\% & 84.37\% & 12.48\% & 82.65\% & 79.57\% & 11.74\% & 10.4\% & \textbf{94.11\%} & 9.4\% \\
        & & & 25\% & 
            14.54\% & 84.74\% & 97.34\% & 13.17\% & 85.88\% & 97.25\% & 26.2\% & 24.65\% & \textbf{99.25\%} & 24.8\% \\
        & & & 50\% &
            14.74\% & 85.25\% & 97.54\% & 13.45\% & 86.54\% & 98.0\% & 50.71\% & 49.28\% & \textbf{99.6\%} & 49.91\% \\
        \cline{2-14}
         & \multirow{3}{*}{98.33\%} & \multirow{3}{*}{25\%} & 10\% & 
            15.92\% & 35.25\% & 34.44\% & 5.82\% & 30.52\% & 22.76\% & 11.04\% & 9.72\% & \textbf{39.29\%} & 4.54\% \\
        & & & 25\% & 
            16.06\% & 83.39\% & 90.04\% & 5.98\% & 86.84\% & 81.42\% & 25.4\% & 25.13\% & \textbf{96.89\%} & 13.06\% \\
        & & & 50\% & 
            16.32\% & 83.68\% & \textbf{99.45\%} & 6.27\% & 93.72\% & 93.93\% & 50.6\% & 49.39\% & 99.42\% & 37.04\% \\
        \cline{2-14}
         & \multirow{3}{*}{16.97\%} & \multirow{3}{*}{50\%} & 10\% & 
            16.53\% & 3.35\% & 3.35\% & 3.2\% & \textbf{16.71\%} & 3.35\% & 9.87\% & 9.86\% & 3.35\% & 10.13\% \\
        & & & 25\% & 
            41.55\% & 8.3\% & 8.3\% & 8.28\% & \textbf{41.68\%} & 8.3\% & 24.94\% & 24.78\% & 8.3\% & 25.25\% \\
        & & & 50\% & 
            \textbf{83.06\%} & 16.93\% & 16.93\% & 16.93\% & \textbf{83.06\%} & 16.93\% & 50.08\% & 49.91\% & 16.93\% & 50.02\% \\
        \hline
        \multirow{9}{*}{Character} & \multirow{3}{*}{94.75\%} & \multirow{3}{*}{10\%} & 10\% & 
            33.33\% & 33.33\% & 81.15\% & 29.71\% & 40.57\% & 52.17\% & 2.17\% & 38.4\% & \textbf{87.68\%} & 8.69\% \\
        & & & 25\% & 
            35.50\% & 57.24\% & 97.1\% & 33.33\% & 61.59\% & 86.95\% & 6.52\% & 57.97\% & \textbf{97.82\%} & 23.91\% \\
        & & & 50\% &
            36.95\% & 63.04\% & \textbf{100.0\%} & 33.33\% & 66.66\% & 96.37\% & 19.56\% & 80.43\% & 99.27\% & 57.97\% \\
        \cline{2-14}
         & \multirow{3}{*}{89.73\%} & \multirow{3}{*}{25\%} & 10\% & 
            30.14\% & 14.2\% & 33.33\% & 28.69\% & 13.04\% & 19.13\% & 9.85\% & 7.85\% & \textbf{39.42\%} & 8.98\% \\
        & & & 25\% & 
            39.13\% & 35.36\% & 70.72\% & 37.97\% & 34.78\% & 44.05\% & 26.66\% & 20.0\% & \textbf{95.36\%} & 27.24\% \\
        & & & 50\% & 
            46.08\% & 53.91\% & 98.26\% & 43.47\% & 56.52\% & 83.76\% & 53.62\% & 46.37\% & \textbf{100.0\%} & 52.17\% \\
        \cline{2-14}
         & \multirow{3}{*}{88.39\%} & \multirow{3}{*}{50\%} & 10\% & 
            \textbf{19.97\%} & 0.57\% & 11.57\% & \textbf{19.97\%} & 0.14\% & 8.24\% & 11.43\% & 6.94\% & \textbf{19.97\%} & 10.56\% \\
        & & & 25\% & 
            \textbf{49.63\%} & 1.44\% & 29.66\% & 49.49\% & 1.59\% & 16.06\% & 28.94\% & 19.82\% & \textbf{49.63\%} & 25.03\% \\
        & & & 50\% & 
            91.17\% & 8.82\% & 57.88\% & 89.86\% & 10.13\% & 35.89\% & 56.0\% & 43.99\% & \textbf{95.8\%} & 49.92\% \\
        \hline
        \multirow{9}{*}{FordB} & \multirow{3}{*}{66.61\%} & \multirow{3}{*}{10\%} & 10\% & 
            45.66\% & 9.44\% & 29.13\% & 45.66\% & 9.05\% & 30.31\% & 6.29\% & 9.44\% & \textbf{70.86\%} & 11.41\% \\
        & & & 25\% & 
            48.03\% & 40.94\% & 64.96\% & 48.03\% & 40.55\% & 57.87\% & 17.71\% & 26.77\% & \textbf{92.51\%} & 25.19\% \\
        & & & 50\% &
            48.42\% & 51.57\% & 99.6\% & 48.42\% & 51.57\% & \textbf{99.6\%} & 46.85\% & 53.14\% & 99.21\% & 48.42\% \\
        \cline{2-14}
         & \multirow{3}{*}{59.83\%} & \multirow{3}{*}{25\%} & 10\% & 
            19.49\% & 27.98\% & 19.81\% & 18.86\% & 28.93\% & 28.93\% & 9.9\% & 7.23\% & \textbf{38.52\%} & 9.43\% \\
        & & & 25\% & 
            35.53\% & 46.38\% & 51.41\% & 33.01\% & 46.38\% & 58.17\% & 22.64\% & 22.48\% & \textbf{75.31\%} & 22.95\% \\
        & & & 50\% & 
            47.32\% & 52.67\% & 93.86\% & 46.22\% & 53.77\% & 78.93\% & 49.05\% & 50.78\% & \textbf{95.44\%} & 49.84\% \\
        \cline{2-14}
         & \multirow{3}{*}{49.78\%} & \multirow{3}{*}{50\%} & 10\% & 
            5.34\% & \textbf{14.15\%} & \textbf{14.15\%} & \textbf{14.15\%} & 5.34\% & \textbf{14.15\%} & 10.22\% & 9.11\% & 13.6\% & 9.74\% \\
        & & & 25\% & 
            18.94\% & 30.42\% & 30.42\% & \textbf{30.5\%} & 19.1\% & \textbf{30.5\%} & 25.39\% & 24.92\% & 29.71\% & 25.23\% \\
        & & & 50\% & 
            48.5\% & \textbf{51.41\%} & 50.7\% & \textbf{51.41\%} & 48.5\% & 50.7\% & 50.07\% & 49.84\% & 51.33\% & 49.84\% \\
        \hline
    \end{tabular}
\end{table*}

Although the process of finding possible mislabeled data can be automated, it is essential to achieve good accuracy when searching for mislabels as they have to be validated manually. Table~\ref{tab:correction} shows the correction ratio assuming that we manually inspected a subset of the data selected according to a ranking of the corresponding debugging approach. The best correction rates are highlighted showing that with the increasing amount of mislabeled data the model performance decreases up to a point where the model is not able to learn the concept anymore and collapses. Intuitively, a model that does not learn the concept should be rather meaningless for the approaches that try to cover the debugging task as they operate directly on the model using the learned concept. 

Surprisingly, by looking at the second-last column the loss-based approach achieved really good correction accuracies, except for the two models that did not learn the concept correctly. One would expect that the more complex methods, using the model to draw detailed conclusions, outperform the loss as they have additional access to more complex computations. Therefore, these results emphasize the use of the training loss for mislabel correction. Against the expectations, the influence-based measurements outperformed the loss, representer, and random method when the model was not able to learn the concept indicating that the influence-based approach does not strongly rely on that. Overall the loss seems to be a good approach concerning the correction ratio but the best correction accuracy does not necessarily lead to the best performance. The mislabels can have more or less impact and it is mandatory to focus on those with the most impact.

\subsection{Experiment 2: Influence of the inspection ratio}
\begin{figure*}[!t]
    \centering
    \subfloat[Anomaly dataset (Quality: 10\% mislabeled)]{\includegraphics[width=0.45\linewidth]{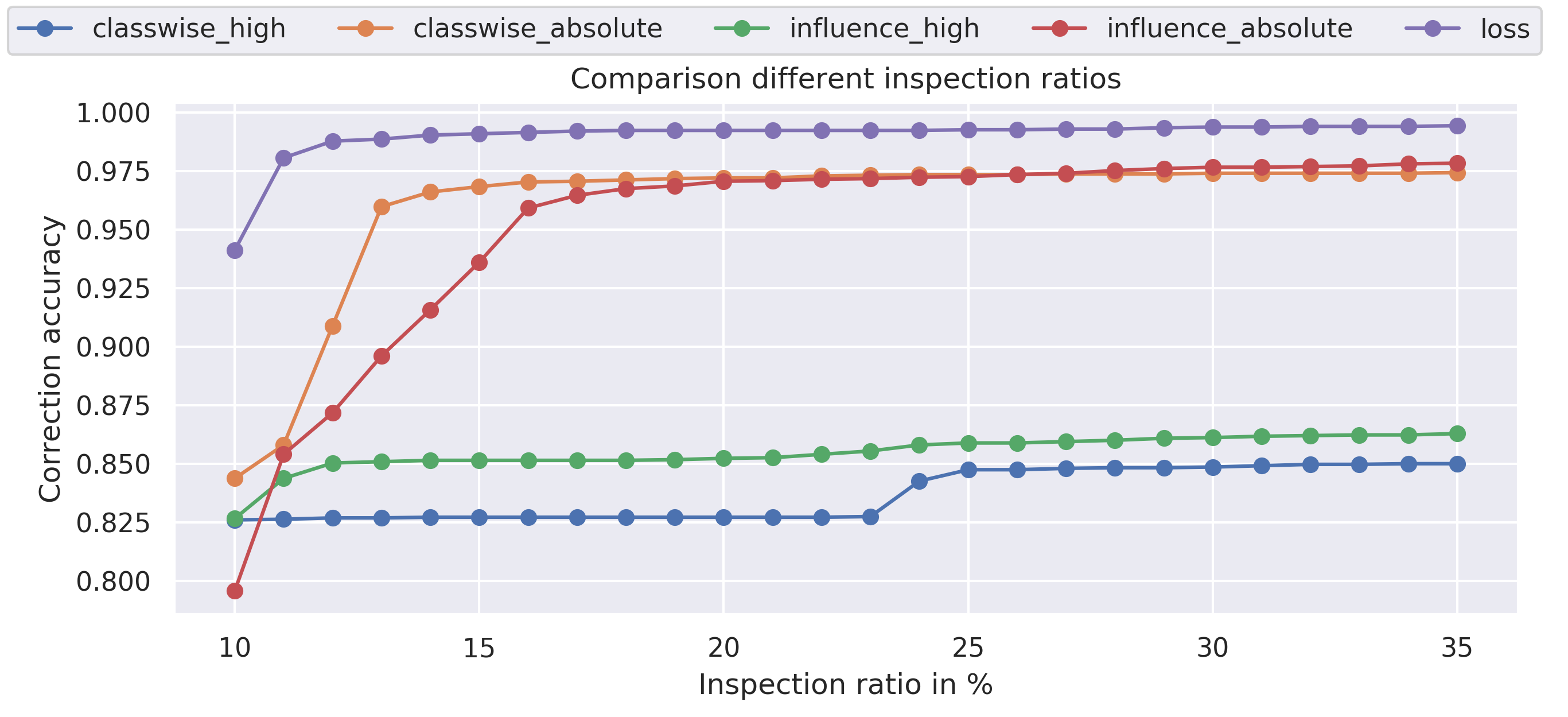}
    \label{fig:ratio10}}
    \subfloat[Anomaly dataset (Quality: 25\% mislabeled)]{\includegraphics[width=0.45\linewidth]{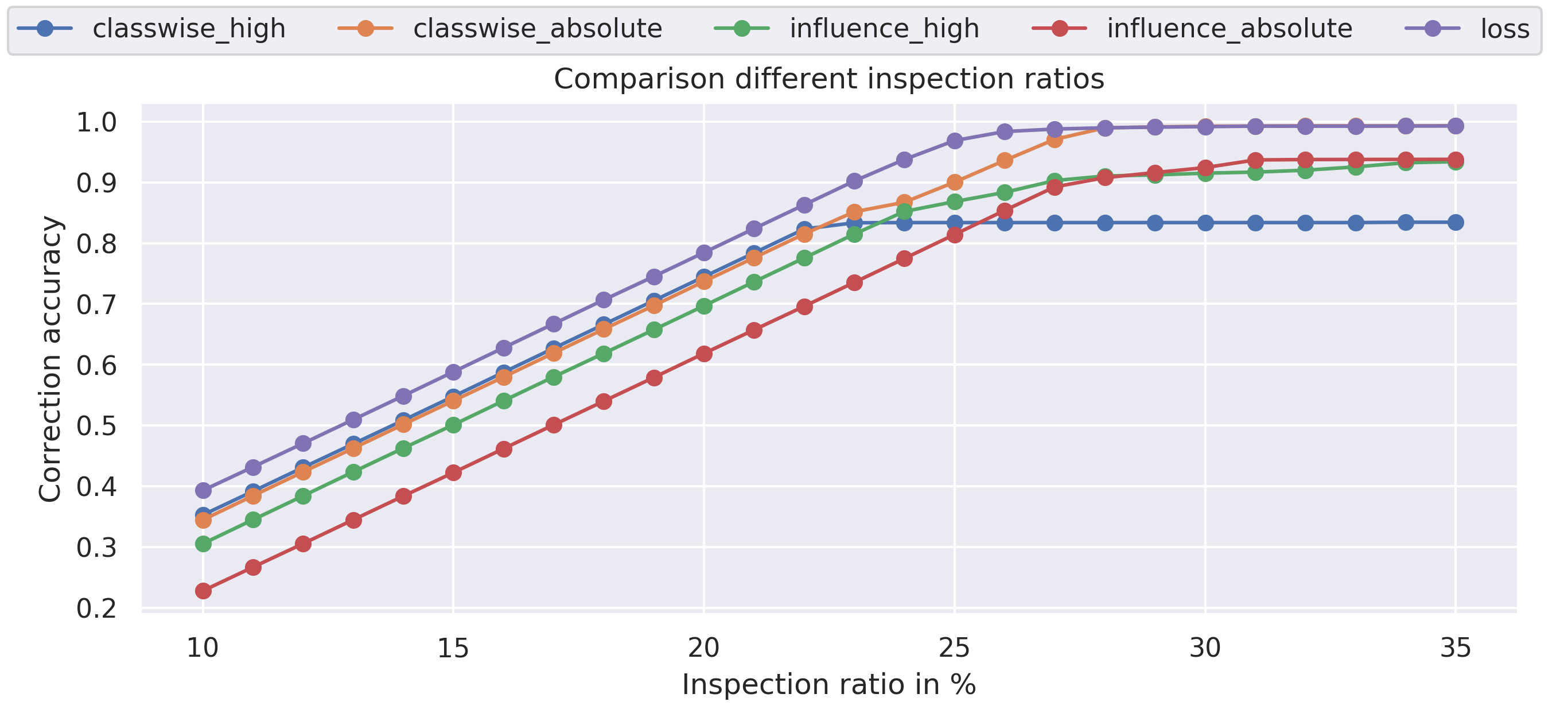}
    \label{fig:ratio25}}

    \subfloat[Character dataset (Quality: 10\% mislabeled )]{\includegraphics[width=0.45\linewidth]{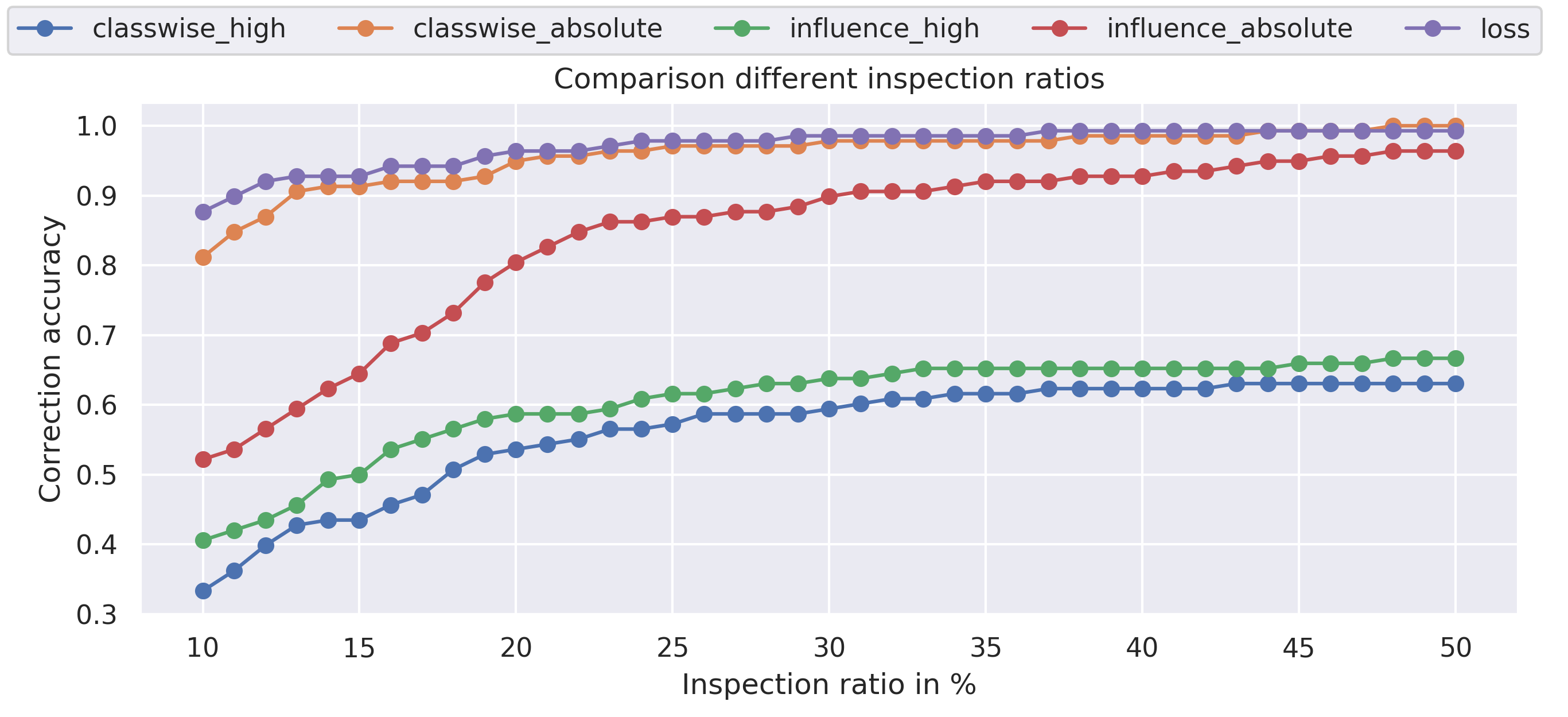}
    \label{fig:ratio10_char}}
    \subfloat[Character dataset (Quality: 25\% mislabeled )]{\includegraphics[width=0.45\linewidth]{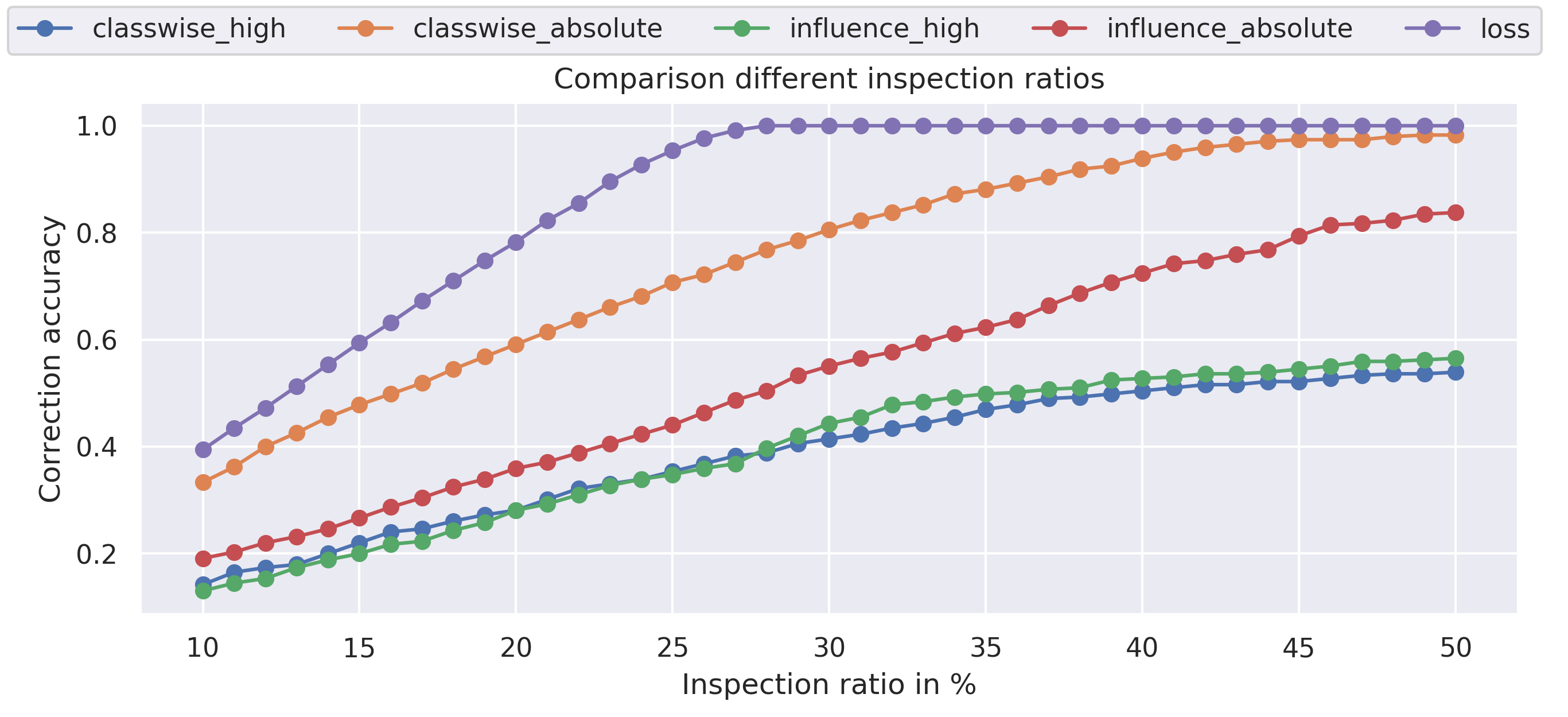}
    \label{fig:ratio25_char}}
    \caption{Different correction accuracies for multiple inspection ratios with a fixed dataset quality.}    
    \label{fig:ratio}
\end{figure*}

We further analyzed the impact of the inspection rate and found out that the gain of a higher inspection rate heavily decreases after a certain point as shown in Figure~\ref{fig:ratio}. The horizontal axis provides the ratio of inspected data after ranking the samples according to the corresponding debugging approach and the vertical axis shows the accuracy of corrected mislabels. In Figure~\ref{fig:ratio10} at 10\% inspected data the correction accuracy should be equal to 0.1 for the random correction and should increase linearly. Both figures do not show all measurements but rather visualize the most successful approaches. The scores in Figure~\ref{fig:ratio10} provide information about the saddle point for the different methods. Also, for the two measurements considering to inspect the most helpful samples, the overall accuracy of the mislabel correction is much lower compared to the other selected methods. Furthermore, the loss outperformed the other methods at any inspection rate. 

In general, Figure~\ref{fig:ratio25} refines the previous results on a different dataset quality. It has to be mentioned that the loss-based method keeps the superior performance. An evaluation of the 50\% mislabeled dataset could not provide meaningful results because the concept was not learned correctly by the model. For a complete analysis and to avoid that the previous finding is related to the properties of the anomaly dataset, the same figures were created for the character dataset because of the diversity of the data and the classification task. In addition, Figure~\ref{fig:ratio} shows the correction accuracies for the character trajectory datasets which reflects that the behavior for the approaches is similar to the results presented for the anomaly dataset. 

\subsection{Experiment 3: Analyzing the score of the correction approaches}
\begin{figure}[!t]
    \centering
    \includegraphics[width=1.0\linewidth]{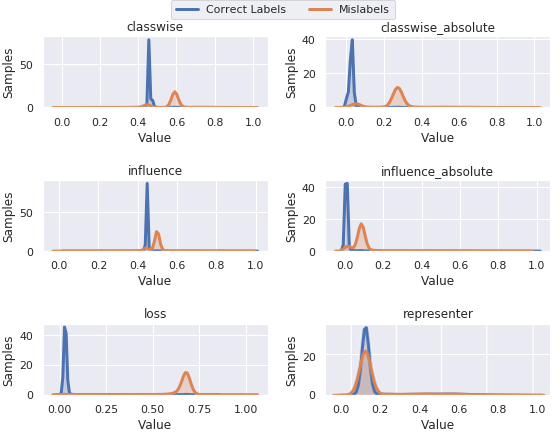}    
    \caption{Normalized distribution of the different correction approaches for the anomaly dataset (Quality: 10\% mislabeled.}
    \label{fig:sampleDistribution}
\end{figure}

\begin{figure}[!t]
    \centering
    \subfloat[classwise]{\includegraphics[width=0.9\linewidth]{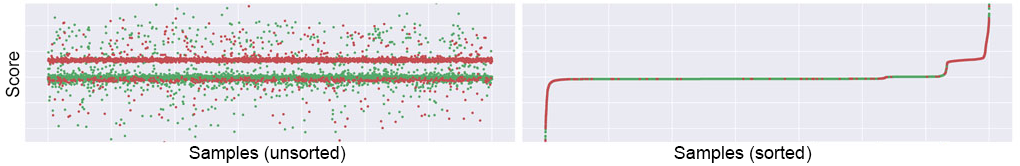}
    \label{fig:sampleRanking_classwise}}
    
    \subfloat[classwise\_absolute]{\includegraphics[width=0.9\linewidth]{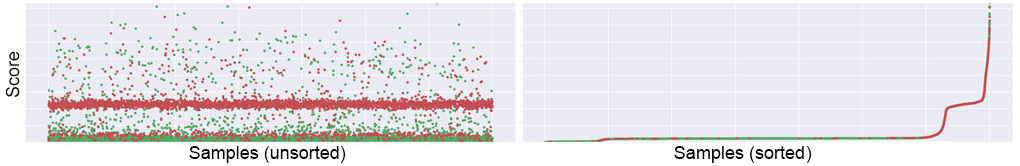}
    \label{fig:sampleRanking_classwise-absolute}}
    
    \subfloat[influence]{\includegraphics[width=0.9\linewidth]{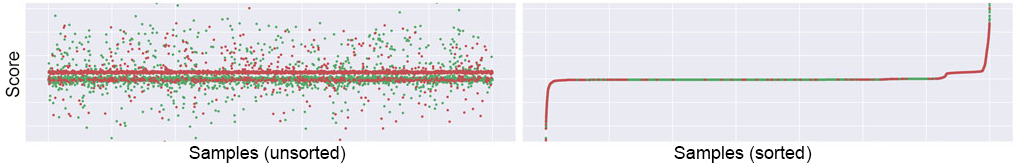}
    \label{fig:sampleRanking_influence}}
    
    \subfloat[influence\_absolute]{\includegraphics[width=0.9\linewidth]{images/anomaly_new_imporances_small_classwise.png}
    \label{fig:sampleRanking_influence-absolute}}
    
    \subfloat[loss]{\includegraphics[width=0.9\linewidth]{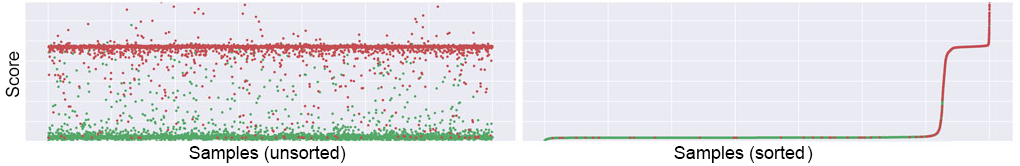}
    \label{fig:sampleRanking_loss}}
    
    \subfloat[representer]{\includegraphics[width=0.9\linewidth]{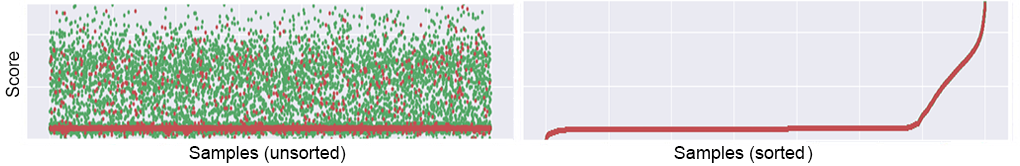}
    \label{fig:sampleRanking_representer}}
    
    \caption{Left column shows the unsorted scores for one of the anomaly datasets (Quality: 10\% mislabeled).}    
    \label{fig:sampleRanking}
\end{figure}

To understand the performance differences a more detailed look into the distribution and the computed values is mandatory. In Figure~\ref{fig:sampleDistribution} the distribution of these values shows that for some methods the distribution highlights the two classes. E.g. the loss-based values show a clear separation of the correct labels and the mislabels. In contrast to that, the representer-based values do not separate the data in such a manner. The same holds for the 'influence absolute' values. Besides those two methods, all other methods provide a very good separation of the data in the distribution plot. Although these plots of the distribution provide a rough understanding of the values more detailed inspection is provided in the following paragraph.

To better align the findings of the distribution plot we visualized the scores for each sample in the anomaly datasets (Quality: 10\% mislabeled) in Figure~\ref{fig:sampleRanking}. The right column shows the sorted scores which were used for the experiments and provide a better overview of the separation of the labels. 

Figure~\ref{fig:sampleRanking_classwise} shows the scores for the classwise measurement in an unsorted (left) and sorted (right) manner indicating that selecting the lowest or highest scores can lead to a good mislabel correction. The high values correspond to the helpful whereas the low are harmful samples and it is possible to improve the quality of those. Figure~\ref{fig:sampleRanking_classwise-absolute} shows the absolute values of this measurement and therefore it is not possible to differentiate between helpful and harmful resulting in a single influence value indicating only the importance concerning the classification.

The approaches shown in Figure~\ref{fig:sampleRanking_influence} and Figure~\ref{fig:sampleRanking_influence-absolute} do not compute the influence separate for each class. This can change the scores for some samples. Especially, if samples are more important for a specific class this measurement does not capture this property. 

In Figure~\ref{fig:sampleRanking_loss} an almost perfect separation provided by the loss-based approach is shown. The loss value for the mislabels is very high compared to the correct-labeled samples and selecting the samples with a high loss indicates to be a very good measurement when the learned concept is meaningful.

Finally, Figure~\ref{fig:sampleRanking_representer} shows the representer values. The plot on the left side maybe leads to the conclusion that the mislabels have lower scores but inspecting the sorted values proves that this is not the case.

\subsection{Experiment 4: Identification differences -- sample ranking}
\begin{table*}[!t]
\tiny
    \centering
    \caption{Detected mislabels for the best combinations. The first row of each setup highlights the best performance without any combination and the following the best combined approaches.}
    \label{tab:correction_combined}
    \begin{tabular}{|c|c|c|c||c|c|c|c|c|c||c|c||c|}
    \hline
        \multirow{2}{*}{\textbf{Dataset}} & \multirow{2}{*}{\textbf{Mislabeled}} & \multirow{2}{*}{\textbf{Inspected}} & \multirow{2}{*}{\textbf{Corrected}} & \multicolumn{6}{c||}{\textbf{Influence-based~\cite{koh2017understanding}}} & \multicolumn{2}{c||}{\textbf{Representer theorem~\cite{yeh2018representer}}} & \multirow{2}{*}{\textbf{Loss}} \\
        & & & & \textbf{classwise low} & \textbf{classwise high} & \textbf{classwise absolute} & 
        \textbf{low} & \textbf{high} & \textbf{absolute} & \textbf{low} & \textbf{high} & \\
        \hline
        \multirow{12}{*}{Anomaly} & \multirow{4}{*}{10\%} & \multirow{4}{*}{10\%} & 94.11\% &
        - & - & - & - & - & - & - & - & X\\
        \cline{4-13}
        & & & \textbf{94.25\%} & X & - & - & - & - & - & - & - & X\\
        & & & 94.22\% & - & - & - & X & - & - & - & - & X\\
        & & & 94.11\% & - & - & - & - & - & - & X & X & X\\
        \cline{2-13}
         & \multirow{4}{*}{25\%} & \multirow{4}{*}{25\%} & 96.89\% &
        - & - & - & - & - & - & - & - & X\\
        \cline{4-13}
        & & & \textbf{96.94\%} & X & - & - & - & X & - & - & - & X\\
        & & & 96.89\% & - & - & - & X & X & - & - & - & X\\
        & & & 96.89\% & - & - & - & - & - & - & X & X & X\\
        \cline{2-13}
         & \multirow{4}{*}{50\%} & \multirow{4}{*}{50\%} & \textbf{83.06\%} &
        - & - & - & - & X & - & - & - & -\\
        \cline{4-13}
        & & & 83.06\% & - & - & - & - & X & - & - & - & X\\
        & & & 83.06\% & - & - & - & - & X & - & - & X & -\\
        & & & 83.06\% & X & - & - & - & - & - & - & - & X\\
        \hline
        \multirow{12}{*}{Character} & \multirow{4}{*}{10\%} & \multirow{4}{*}{10\%} & 87.68\% &
        - & - & - & - & - & - & - & - & X\\
        \cline{4-13}
        & & & \textbf{89.85\%} & X & - & - & - & - & - & - & - & X\\
        & & & 89.85\% & - & - & X & - & - & X & - & X & X\\
        & & & 88.4\% & - & - & - & X & - & - & - & - & X\\
        \cline{2-13}
         & \multirow{4}{*}{25\%} & \multirow{4}{*}{25\%} & 95.36\% &
        - & - & - & - & - & - & - & - & X\\
        \cline{4-13}
        & & & \textbf{96.81\%} & X & - & - & - & - & - & - & - & X\\
        & & & 96.23\% & - & - & - & X & - & - & - & - & X\\
        & & & 95.36\% & - & - & - & - & - & - & X & X & X\\
        \cline{2-13}
         & \multirow{4}{*}{50\%} & \multirow{4}{*}{50\%} & 95.8\% &
        - & - & - & - & - & - & - & - & X\\
        \cline{4-13}
        & & & \textbf{96.52\%} & X & - & - & - & - & - & - & - & X\\
        & & & 96.09\% & X & X & - & - & - & - & - & - & X\\
        & & & 95.8\% & - & - & - & - & - & - & X & X & X\\
        \hline
    \end{tabular}
\end{table*}

Although we showed that some methods separate the data better, we decided to have a more detailed look at the samples that are not detected and the samples that are only detected by a specific method because not every sample has the same weight towards the classification accuracy. This is especially of interest when it comes to the classification performance rather than the correction accuracy. In theory, it is a good practice to aim for the highest mislabel correction rate but this does not mandatory result in the best possible classifier. Therefore, a more detailed inspection of the different detected samples followed by an accuracy evaluation can provide a better understanding of the results as this could favor the influence functions~\cite{koh2017understanding} and representer point~\cite{yeh2018representer} performances.

\begin{figure}[!t]
    \centering
    \includegraphics[width=0.9\linewidth]{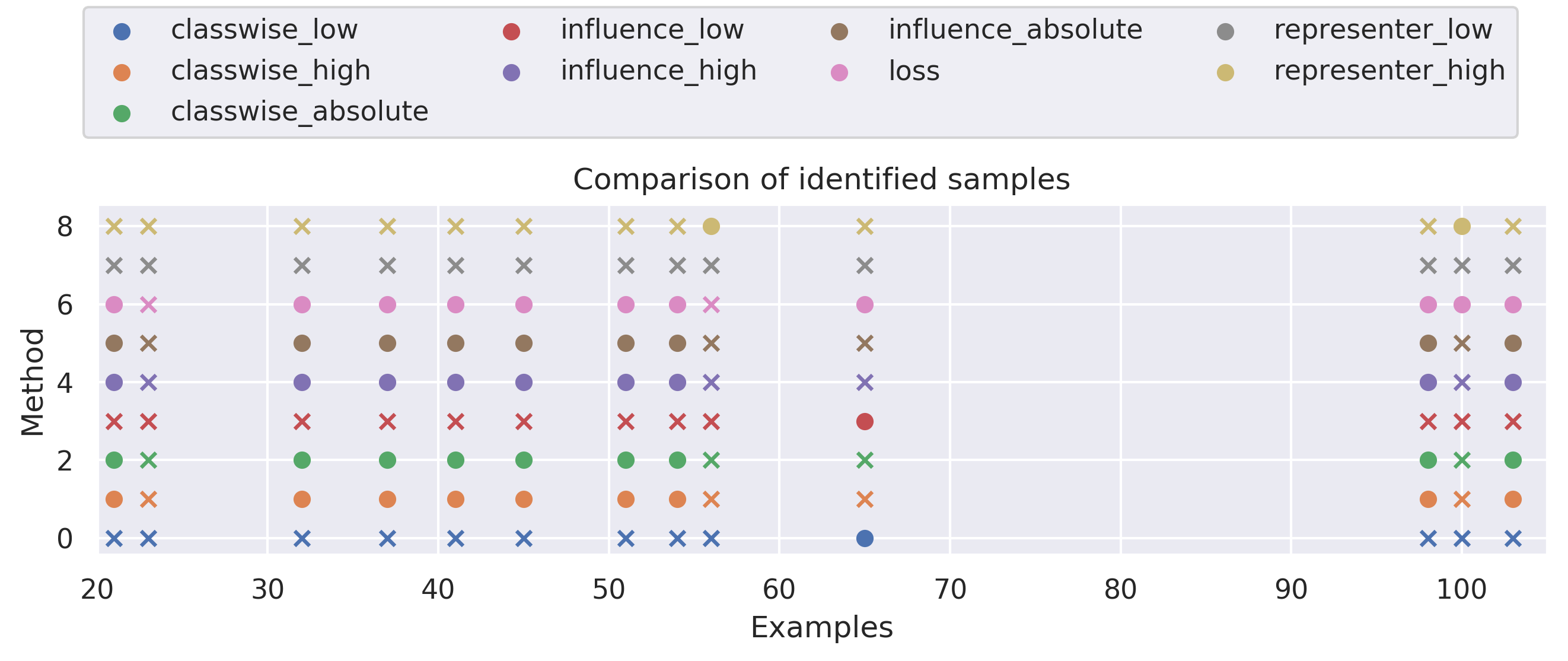}
    \caption{First 100 samples of the anomaly dataset (Quality: 10\% mislabeled). Dots indicate detected and crosses undetected mislabels.}
    \label{fig:sampleSlice}
\end{figure}

As shown in Figure~\ref{fig:sampleSlice} the approaches detect different mislabels and a combination of the approaches could provide better correction results. For example, the representer method only detects two out of the 13 mislabels but one of these is not detected by any other methods. Especially, the loss-based method which detects 11 out of the 13 shown label flips was not able to detect this sample. 

\subsection{Experiment 5: Combining correction approaches}
Concerning previous findings, a combination of the approaches could lead to even better results. To combine the methods, we normalized the ranking scores to make it possible to compare them linearly. Although this combination approach is very simple the results show the capabilities of a combination.

Table~\ref{tab:correction_combined} presents the results for some selected combinations. The results refine the findings that the loss, as a baseline, is really good, and only in the case where the model did not learn the concept, the loss is significantly worse than the other approaches. Also, it shows that the combined methods can reach a very stable performance for the 50\% mislabeled anomaly dataset. The results for the character trajectories dataset are similar to those of the anomaly dataset. Besides, the combinations with the loss perform well even for the 50\% mislabel due to the correctly learned concept. 

Furthermore, these experiments emphasize that a combination can improve the correction accuracy and improve the robustness compared to the use of a single measurement. Nevertheless, drawbacks exist addressing the computational effort and the robustness as shown in Table~\ref{tab:correction}. Some methods are not as reliable as the results of the loss and using them can decrease the performance as well.

\subsection{Experiment 6: Additional time consumption}
\begin{figure}[!t]
    \centering
    \includegraphics[width=0.9\linewidth]{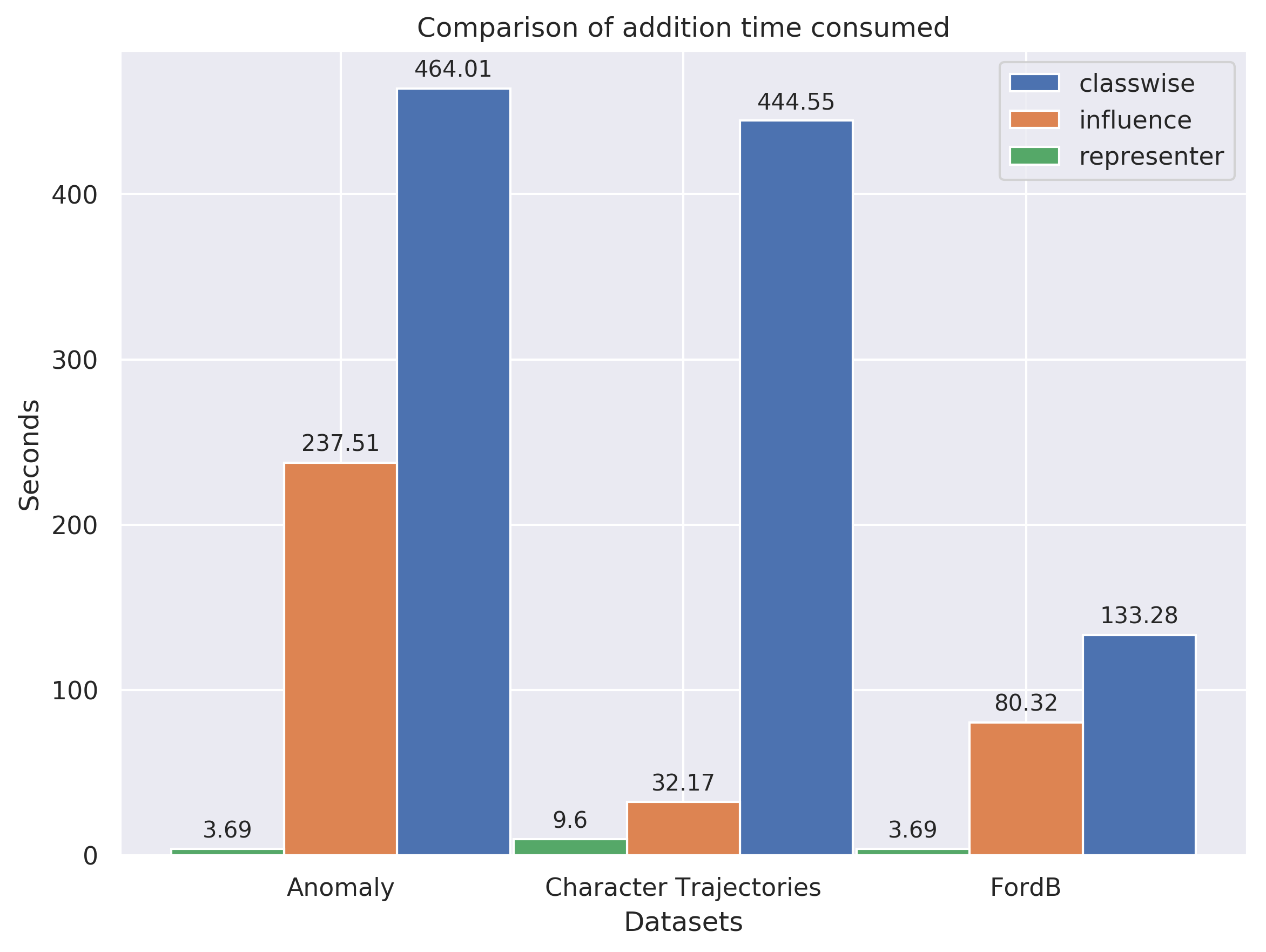}
    \caption{Additional computation time excluding any measurement that can be done during the evaluation process.}
    \label{fig:time}
\end{figure}

In contrast to the loss-based approach, the others need additional computation time. The training loss can be collected during the evaluation process without a significant slowdown. The influence function~\cite{koh2017understanding} needs an already trained model and the execution of this method is extremely time-consuming. Especially, the computation of the classwise measurement requires a lot of time. The same holds for the representer-based method~\cite{yeh2018representer}. This method needs additional training to learn the representation to compute the representer value based on the pre-softmax activations. In contrast to the influence-based methods, this additional training is class independent and depends on representation size. 

The time consumption is visualized in Figure~\ref{fig:time} and the loss is excluded. As for the other approaches, the representer method has very low computational extra time. The computational effort for the influence strongly depends on the dataset size. Also, the computational effort for the classwise measurement suffers from the number of different classes. A comparison of the datasets showed that for the anomaly and FordB dataset the computation time for the classwise measurement increased about 40\% for the FordB dataset and 50\% for the anomaly dataset as both have two classes. The Character trajectories dataset has 20 classes and therefore the increase in additional time is much higher.

\subsection{Experiment 7: Detailed sample analysis}
\begin{figure}[!t]
    \centering
    \includegraphics[width=1.0\linewidth]{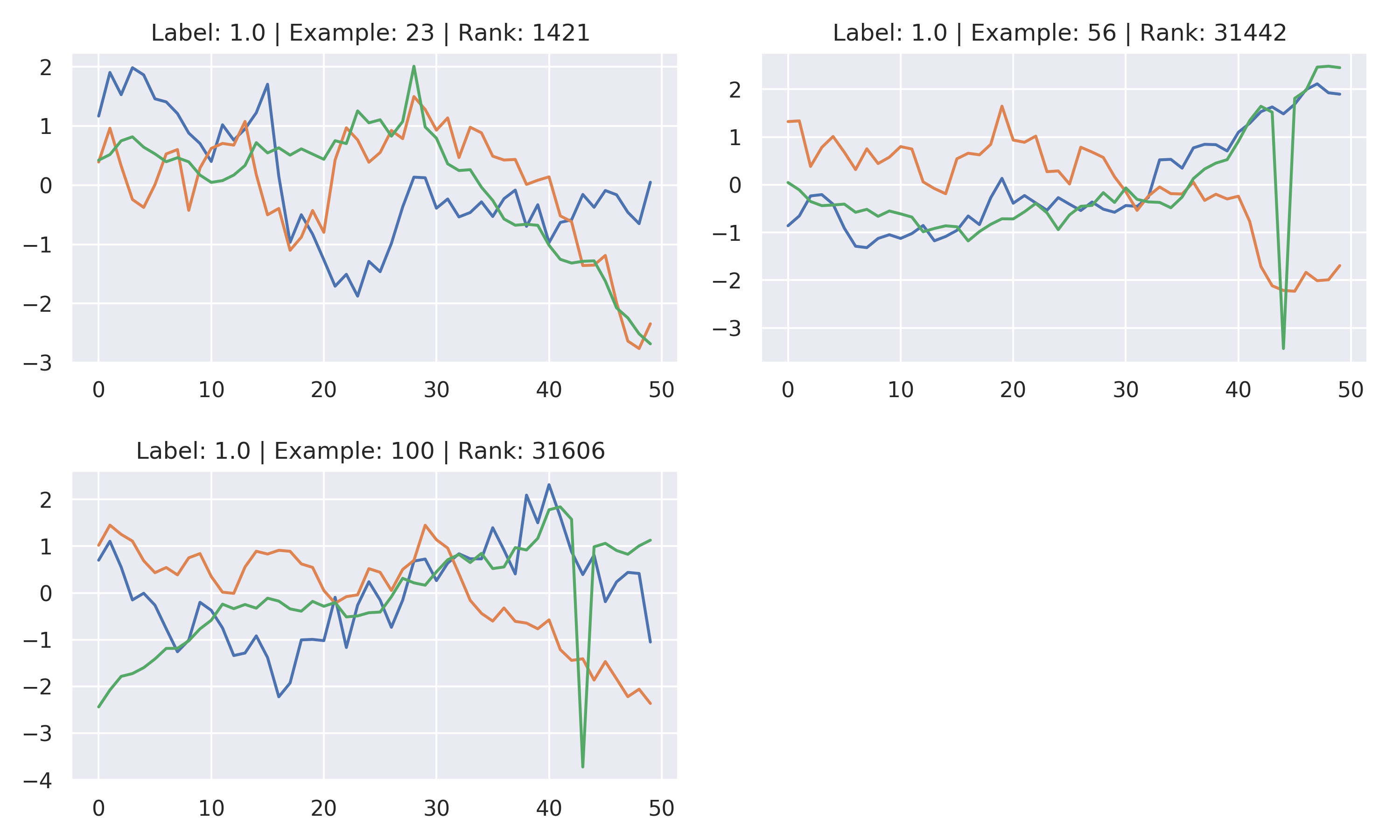}
    \caption{Three selected samples for the loss based correction. All samples are anomalies within the ground truth but their labels were flipped during the training. Only sample 100 was successfully identified as a mislabel.}
    \label{fig:sampleSet}
\end{figure}

\begin{figure*}[!t]
    \centering
    \subfloat[Samples with the lowest loss (labeled as no anomaly and anomaly)]{\includegraphics[width=0.3\linewidth]{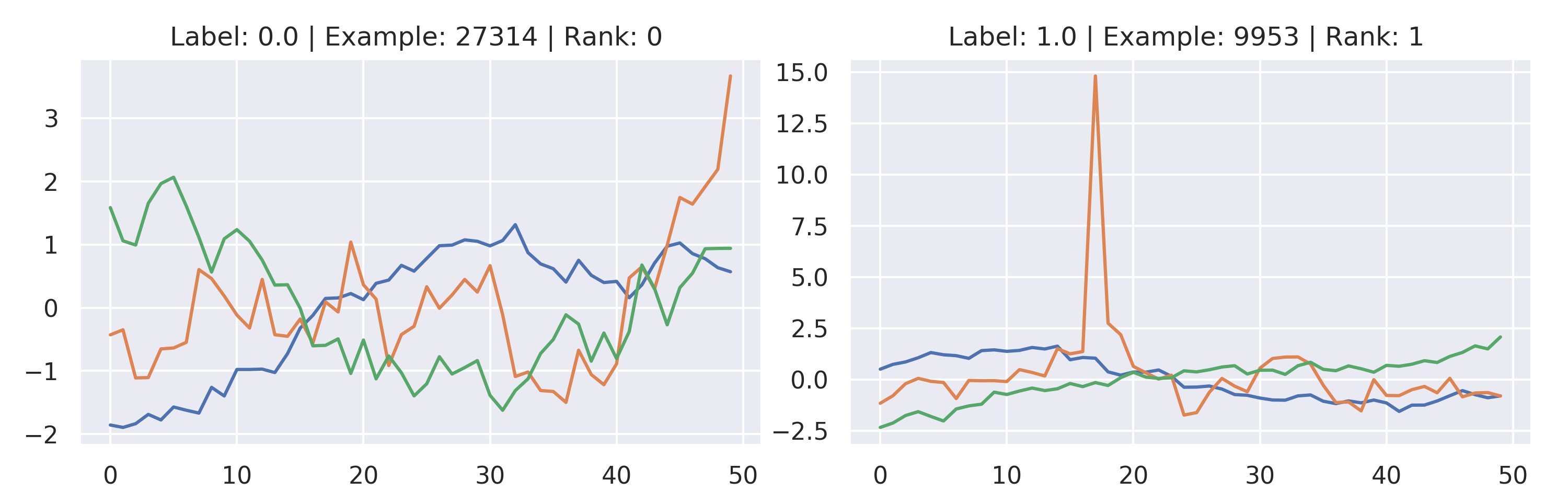}
    \label{fig:lossLow}}
    \subfloat[Samples with the highest loss (both mislabeled as anomalies)]{\includegraphics[width=0.3\linewidth]{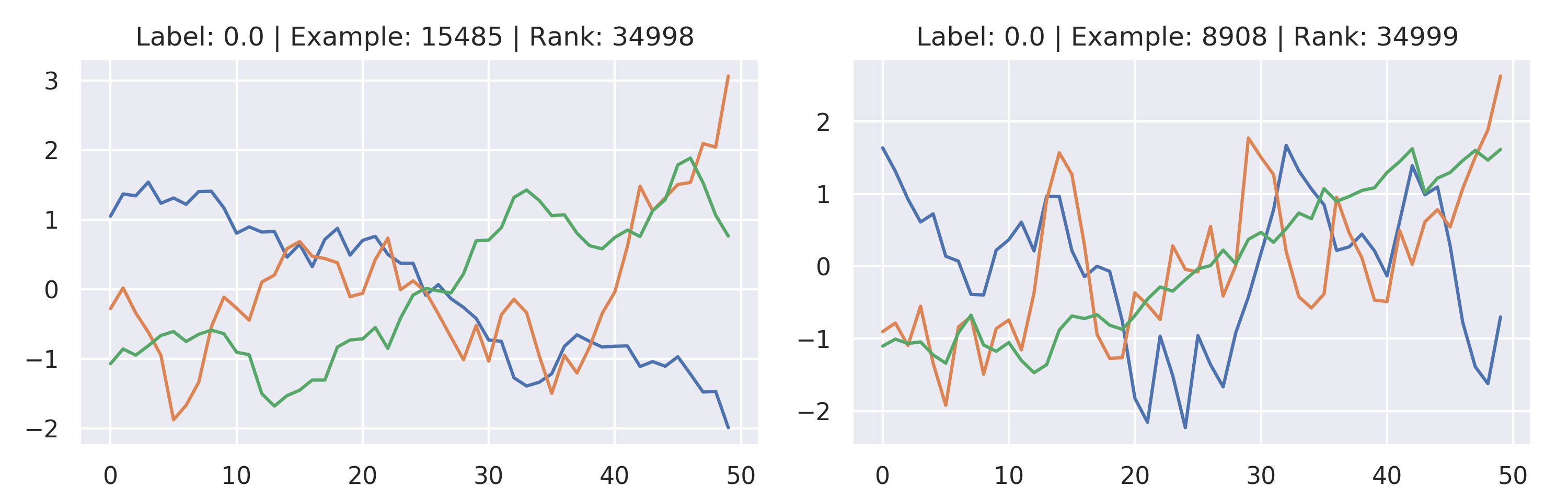}
    \label{fig:lossHigh}}
    
    \subfloat[Harmful samples with negative influence value (both mislabeled as no anomalies)]{\includegraphics[width=0.3\linewidth]{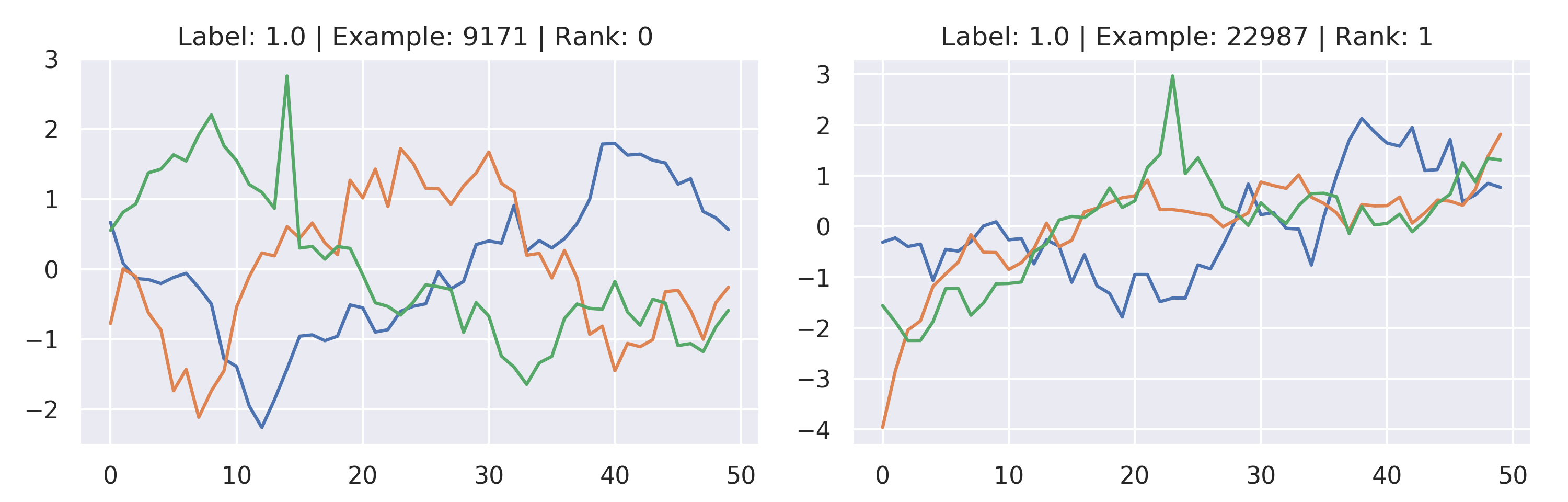}
    \label{fig:influenceNeg}}
    \subfloat[Helpful samples with positive influence value (both labeled as anomalies)]{\includegraphics[width=0.3\linewidth]{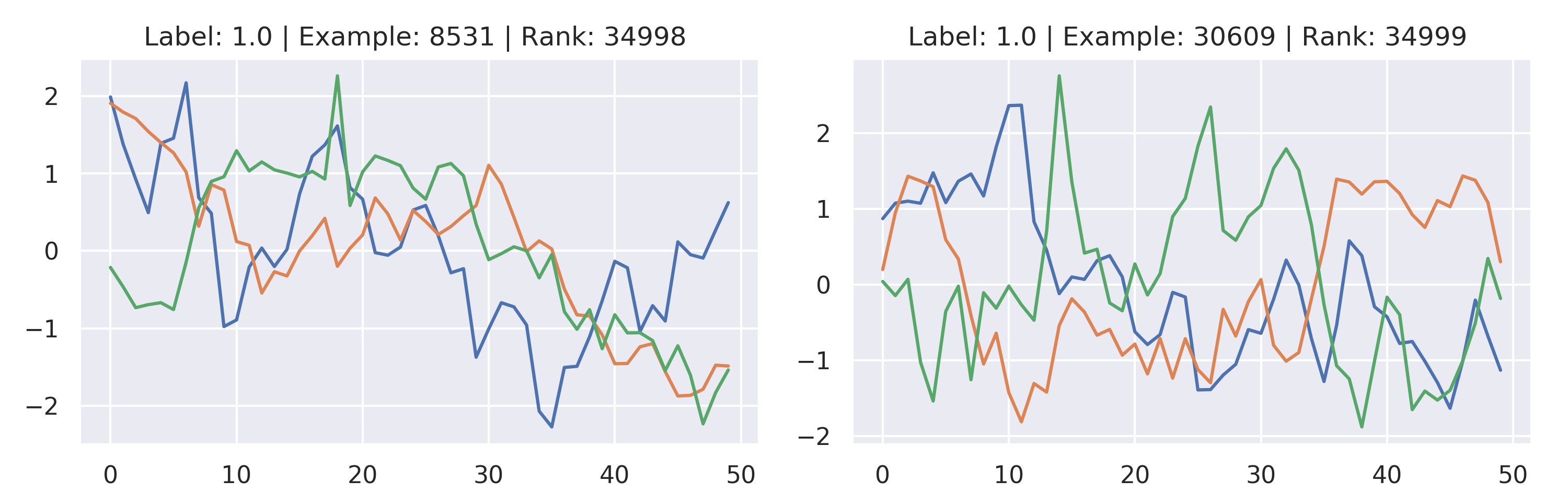}
    \label{fig:influencePos}}
    \subfloat[Samples with low absolute influence value, low impact (both labeled as no anomalies)]{\includegraphics[width=0.3\linewidth]{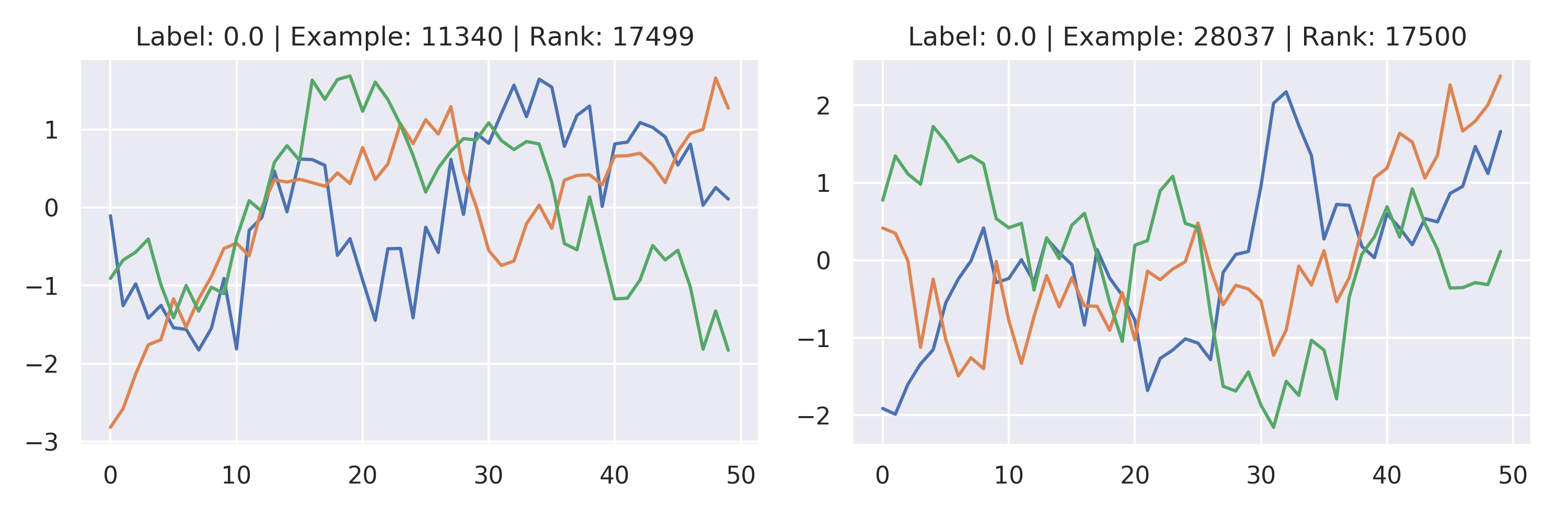}
    \label{fig:influenceNeu}}
    
    \subfloat[Samples with the lowest absolute representer value (both labeled as no anomalies)]{\includegraphics[width=0.3\linewidth]{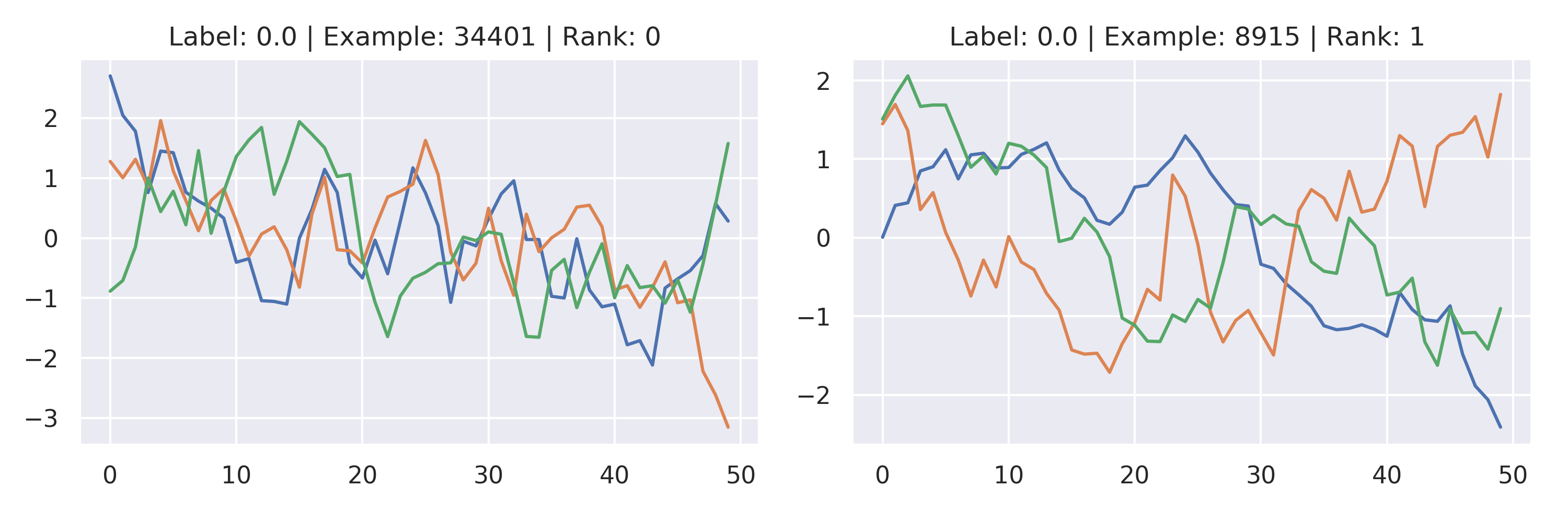}
    \label{fig:representerLow}}
    \subfloat[Samples with the highest absolute representer value (labeled as anomaly and no anomaly)]{\includegraphics[width=0.3\linewidth]{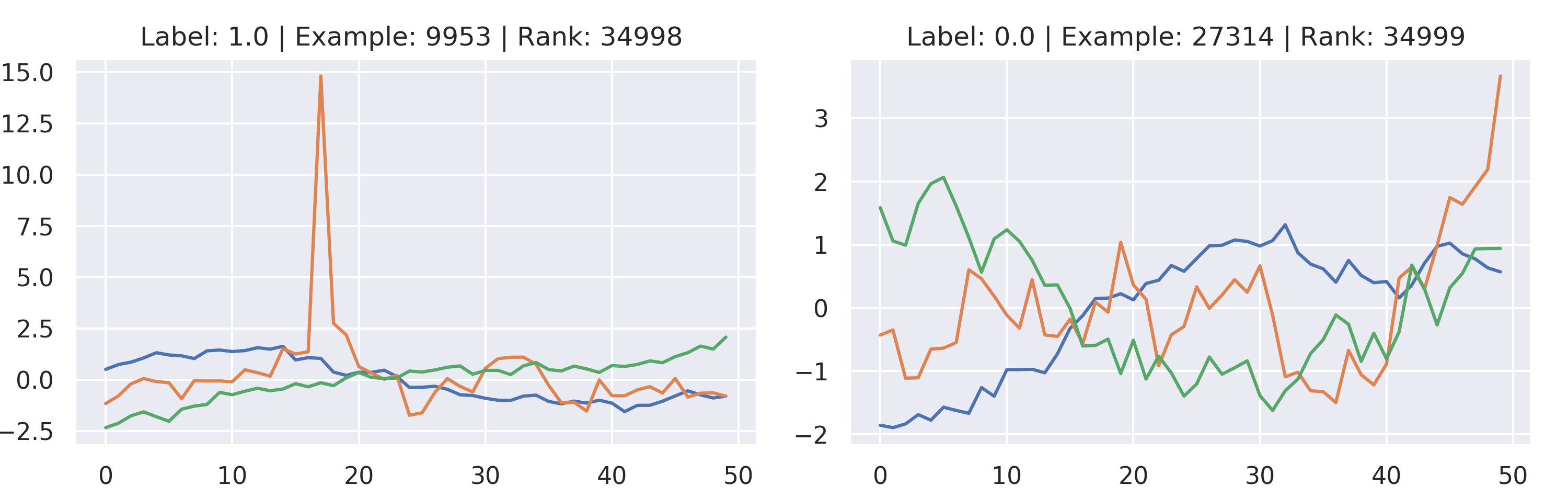}
    \label{fig:representerHigh}}
    \caption{Different selected samples and their scores based on the used approach.}    
    \label{fig:SamplesLowHigh}
\end{figure*}

\begin{figure}[!t]
    \centering
    \subfloat[Mislabels found by loss]{\includegraphics[width=0.9\linewidth]{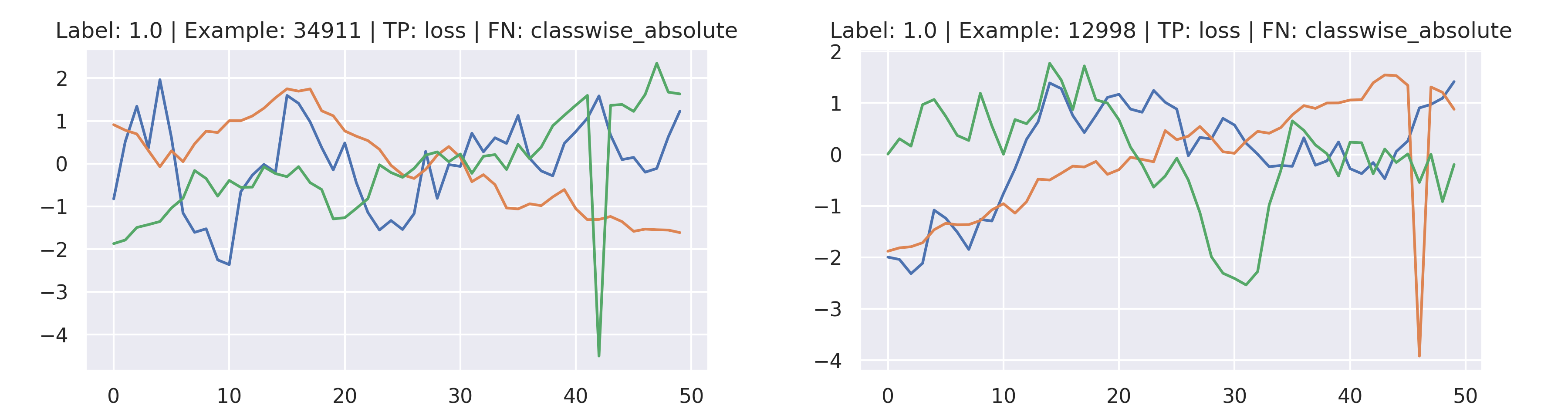}
    \label{fig:lossFound}}
    
    \subfloat[Mislabels found by influence]{\includegraphics[width=0.9\linewidth]{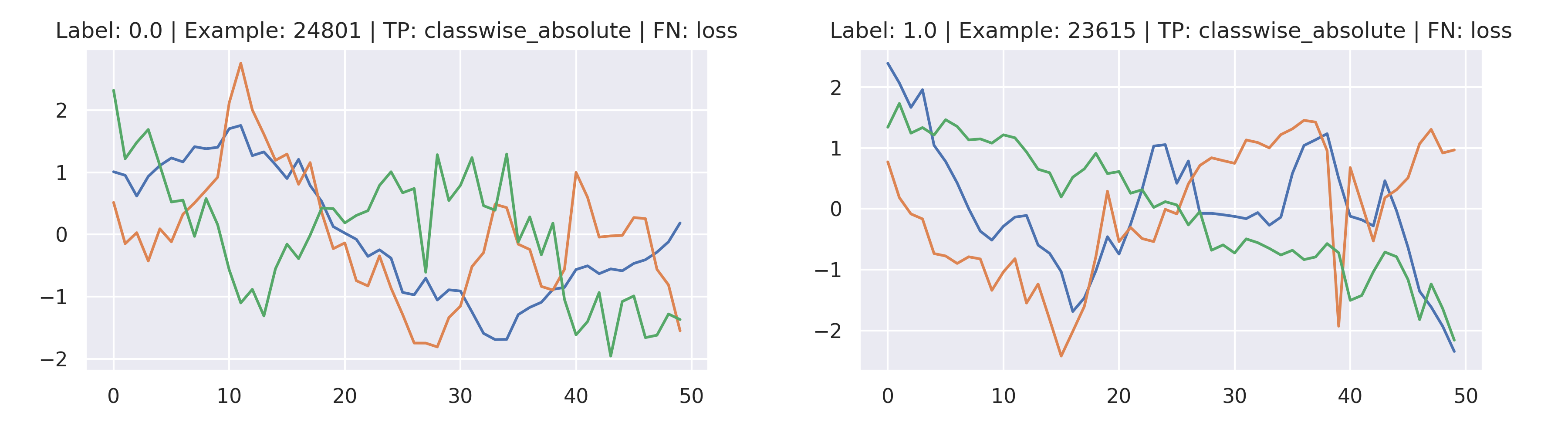}
    \label{fig:influenceFound}}
    \caption{Mislabeled samples that are only found either by the loss or influence approach.}    
    \label{fig:lossInfluenceComparison}
\end{figure}

There are two important questions during the dataset debugging: Why are some samples harder to identify compared to the majority of samples? How do these samples look like and do they provide any information concerning the learned concept? Answering these questions or inspecting these samples can help to interpret the model.

According, to our previous findings that not all samples are similarly easy to find we investigated the difficulty and properties of the samples. It has to be highlighted, that these results are visualized for the anomaly dataset due to the easier interpretability of the problem but could be visualized for the other datasets as well.

In Figure~\ref{fig:sampleSet} three samples of the previously mentioned slice for the loss-based approach are shown. These samples were selected to emphasize the specific properties of the approach. The label shows the correct label whereas one corresponds to the anomaly and zero to the non-anomaly class. Therefore, all samples are classified as anomalies within the ground truth. Only the last sample (second row) was found by inspecting 10\% of the data as this includes the ranks 31500 to 35000 for the training dataset. The rank reflects the position in the dataset sorted according to a specific measurement e.g. loss. Furthermore, the second example (first row, right) was close to the threshold, and increasing the amount of inspected data to 12\% would be sufficient to find this mislabel. Finally, for the first example (first row, left), there is an ambiguity concerning its ground-truth label as it could either be a true mislabel or the model was not able to capture the precise concept of point anomaly concerning the less dominant peak. 

According to the dataset creation process, the sample has the correct ground-truth label highlighting that when it comes to the interpretation and explainability of the model this sample shows that the concept was not precisely learned. With this information, it is possible to include samples related to the missing concept parts or weight these kinds of samples to adjust the learned concept to cover the complete task.

This means, that based on the ranking we can try to understand the learned concept and the dataset quality. Both can help to provide an understanding of the model to improve it. Also, the corresponding influence score ranks the ambiguous sample at position 25556. This information states that the sample is not relevant to the classifier. This assumption is further validated by Figure~\ref{fig:sampleRanking} where the influence of the sample is zero. Therefore, it is not helping or harming the classifier's performance. The same result is given by the classwise influence score which has rank 23197 and following the same procedure results show that this sample does not contribute much to the classifier. Finally, to provide the complete information for that sample, the score for the representer which ranks the sample at rank 4079 was checked and refines the assumption as well. 

Using the information above it is now possible to understand the mislabel as this sample was not important for the classifier. To adjust the classifier to detect peaks like that it is mandatory to increase the importance of these kinds of samples.

After the first conclusions based on the ambiguous sample, we decided to further analyze this direction. Therefore, Figure~\ref{fig:SamplesLowHigh} provides information about the importance of the samples with the highest and lowest scores. Starting with Figure~\ref{fig:lossLow} the two samples with the lowest loss are shown. These samples visualize two pretty good samples for the anomaly detection task. Their loss highlights the learned concept. In contrast to that, Figure~\ref{fig:lossHigh} shows the samples with the highest loss. Important for these two samples is that they were mislabeled. Both had the anomaly label and as the figure shows they should be classified as no anomaly samples. Therefore, their high loss shows that the model correctly learned the concept of anomaly detection. The same plots for the influence are shown in Figure~\ref{fig:influencePos} for the positive influencing samples, Figure~\ref{fig:influenceNeg} for the negative influencing samples and Figure~\ref{fig:influenceNeu} for the least influencing samples. The negative influencing plots show that the classifier works correctly as both are mislabeled samples and the positive influencing and neutral ones are correctly labeled. Finally, Figure~\ref{fig:representerLow} shows the samples with a low representer value and Figure~\ref{fig:representerHigh} the ones with high values. These samples do not include any mislabel. The combination of these insights again emphasizes that including the data and additional debugging methods it is possible to not only detect the mislabeled samples but further show that the concept of the classifier is learned correctly. 

As mentioned early on, the approaches detect different samples. Figure~\ref{fig:lossInfluenceComparison} shows some samples that are found either by the loss based or the influence based method~\cite{koh2017understanding}. For example, the loss-based measurement provides be best mislabel correction rate if the model has a vague understanding of the problem but it does not rank the samples according to their influence. Therefore, it could be that a significant lower mislabel correction accuracy results in superior classification accuracy. Contrary, the influence-based method provides information on how helpful and harmful the samples are but does not maximize the mislabel correction accuracy. 

\subsection{Experiment 8: Model accuracy comparison}
\begin{figure}[!t]
\includegraphics[width=0.9\linewidth]{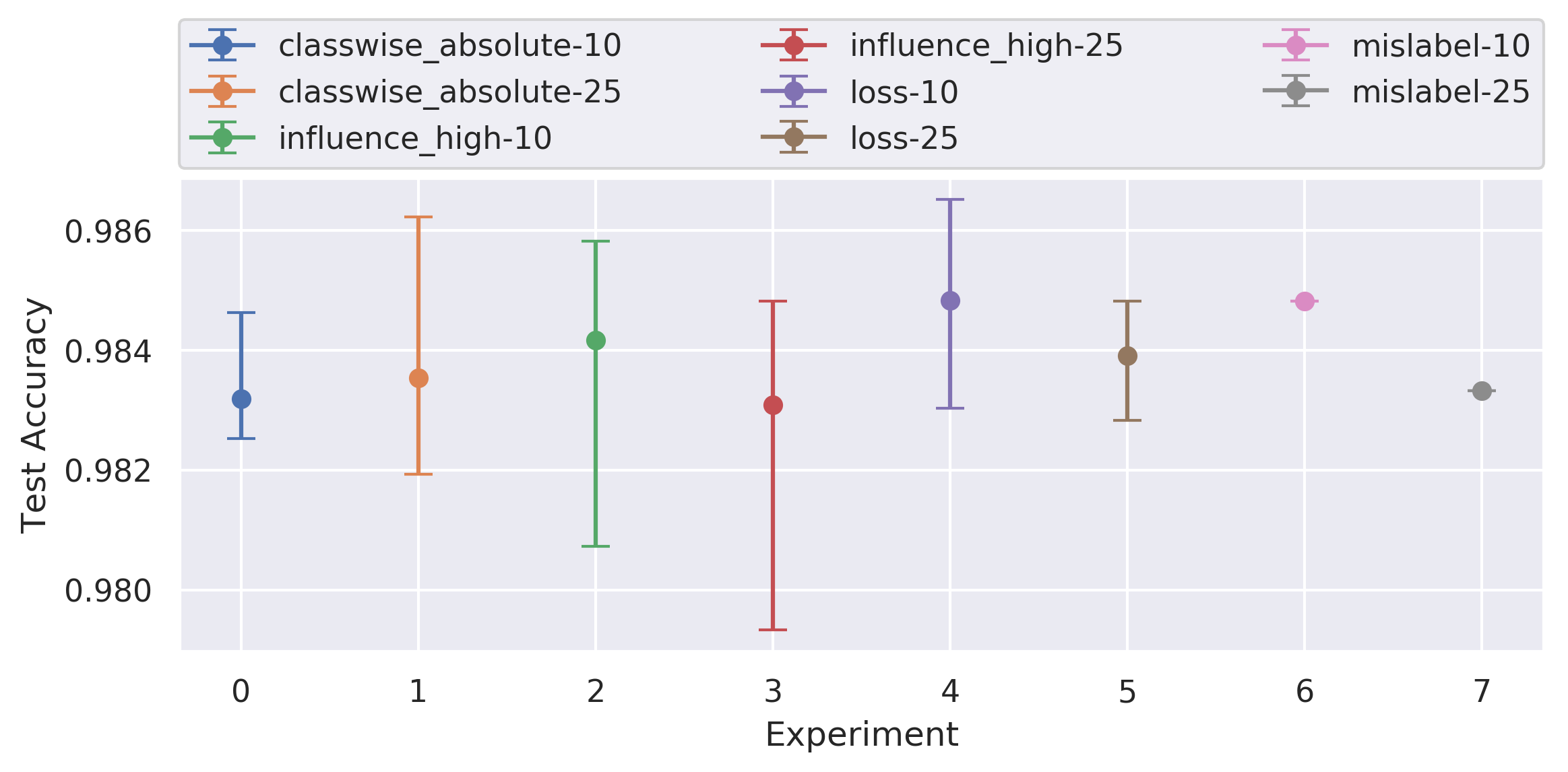}
    \caption{Accuracies of the different models for the anomaly dataset (Quality 10\% and 20\% mislabeled) for the correction task.}    
    \label{fig:accuracy}
\end{figure}

\begin{figure}[!t]
    \centering
    \subfloat[Accuracies for 10\% deletion data.]{\includegraphics[width=0.9\linewidth]{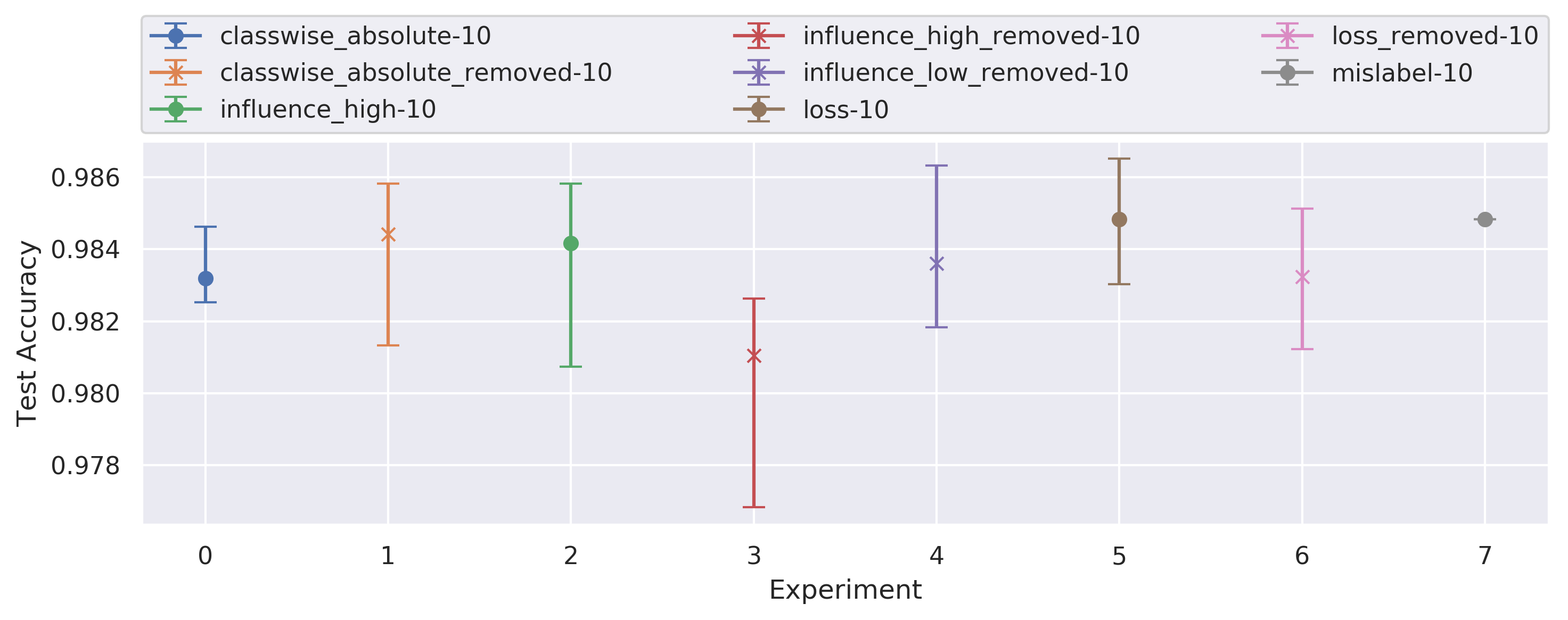}
    \label{fig:accuracy_removed_10}}
    
    \subfloat[Accuracies for 25\% deletion data.]{\includegraphics[width=0.9\linewidth]{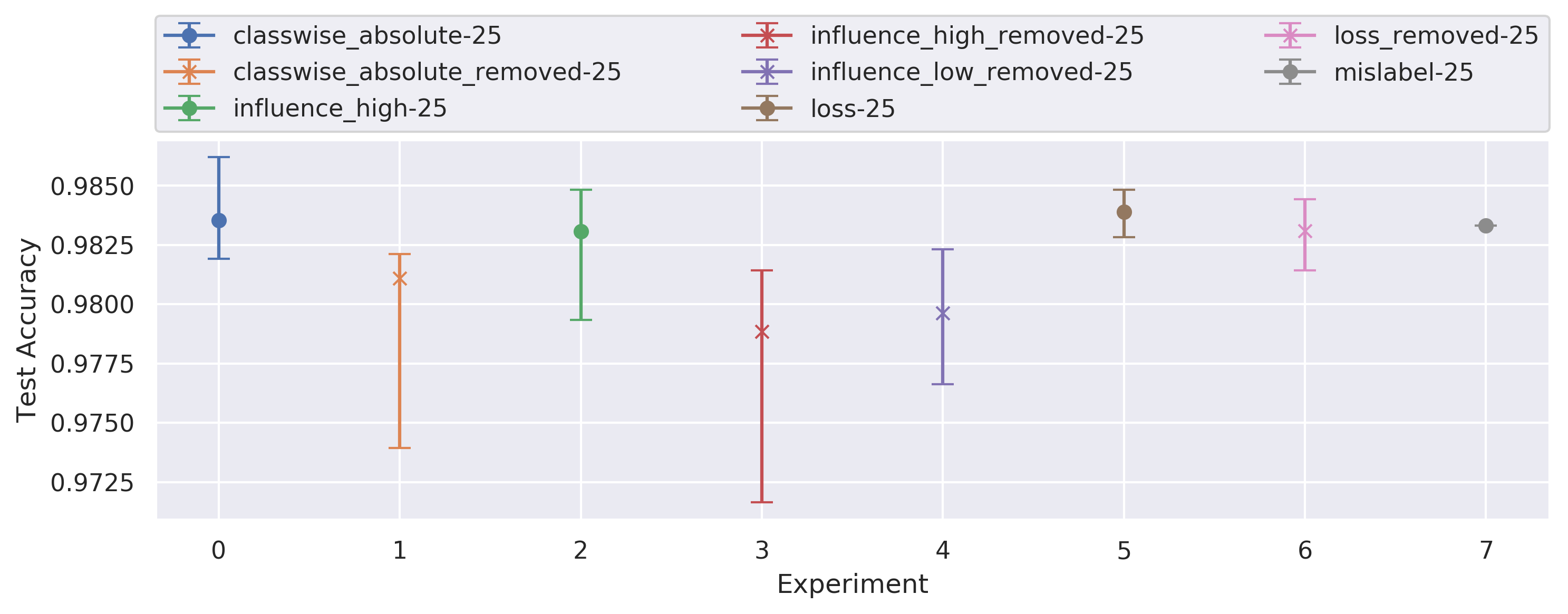}
    \label{fig:accuracy_removed_25}}
    \caption{Accuracies of the different models for anomaly dataset (Quality: 10\% and 20\% mislabeled) for the deletion task.}    
    \label{fig:accuracy_removed}
\end{figure}

To complete the comparison of the methods we present the change in the accuracy for some representative experiments for the anomaly detection dataset. In Figure~\ref{fig:accuracy} it is shown that the accuracy over ten runs for the 10\% mislabeled dataset and the 20\% mislabeled dataset is much better for some approaches and that the variance between the runs is very small concerning the data quality.

Another aspect that is related to the previous analysis is the deletion of a subset based on the measurements. The suggested samples are deleted from the dataset instead of the manual correction which needs time and additional effort. Therefore, the deletion of samples can be executed without human inspection and if the measurement is good it should remove mislabeled data as well as other samples that harm the performance of the classifier. This results in a smaller dataset with improved data quality.

Figure~\ref{fig:accuracy_removed} shows the performances for the mislabel correction compared to the deletion without inspection. In Figure~\ref{fig:accuracy_removed_10} the deletion performed better for the 'classwise absolute' influence computation removing the most influencing samples. Further, the scores for the influence computatio~\cite{koh2017understanding} show that the deletion of samples with low scores improved the accuracy and the deletion of samples with high scores decreased the accuracy reflecting the influence score concerning its definition of helpful and harmful samples. For the loss, we can see that the accuracy drops if we delete the samples. This is especially the case because for the loss-based procedure the correction accuracy is really good and the deletion of the samples just shrinks the data. The results show that except for the loss the accuracies dropped compared to the mislabeled dataset. If a manual inspection is not a valid solution, the deletion of the samples based on the scores do not improve the quality of the data either.

\subsection{Approach comparison}
When it comes to a stable, robust, and effective method to debug mislabels the loss-based approach outperforms the other methods in accuracy and time consumption significantly. The only drawback is that there is no information about the influence of the detected samples as this approach is not used for interpretability. The influence functions have shown to achieve nearly comparable results. Especially, when using the absolute values to check both the harmful and helpful samples the correction rate is stable providing additional influence information. The only drawback is the additional time, especially when the classwise evaluation is used. The representer point was outperformed by a large margin making it not possible to compare it to the superior methods.

\section{Conclusion}
We performed a comprehensive evaluation concerning the topic of automatic mislabel detection and correction. Therefore, we examined multiple experiments and evaluated the performance of two well-known existing methods in the domain of model interpretability. In contrast to the expectations, the loss-based method can handle the mislabel detection task better even though it is a direct measurement and the two already existing methods provide a much deeper understanding of the model. Also, we showed that a combination of the methods can be more robust and lead to even better results. Furthermore, it has to be mentioned that the dataset debugging was only a subtask of the influence and representer approach. Therefore, we presented results that help to interpret the model from a data-based perspective and used different measurements to provide an overview of the models' behavior. We identified the most important samples for the model concerning the different approaches. Finally, we found that the deletion of the suggested mislabeled data does not work better than keeping the mislabeled data.

\section*{Acknowledgements}
This work was supported by the BMBF projects DeFuseNN (Grant 01IW17002) and the ExplAINN (BMBF Grant 01IS19074). We thank all members of the Deep Learning Competence Center at the DFKI for their comments and support.

\bibliographystyle{IEEEtran}
\bibliography{bibliography}

\end{document}